\documentclass{article}

\usepackage{PRIMEarxiv}

\usepackage[utf8]{inputenc} 
\usepackage[T1]{fontenc}    
\usepackage{url}            
\usepackage{booktabs}       
\usepackage{amsfonts}       
\usepackage{nicefrac}       
\usepackage{microtype}      
\usepackage{lipsum}
\usepackage{fancyhdr}       
\usepackage{graphicx}       
\graphicspath{{media/}}     
\usepackage{natbib}
\newtheorem{example}{Example}

\usepackage[colorlinks,bookmarksopen,bookmarksnumbered,citecolor=red,urlcolor=red]{hyperref}
\usepackage[TikZ]{mdframed}
\usepackage{listings}
\usepackage{xcolor}
\usepackage{colortbl}
\lstdefinelanguage{json}{
    basicstyle=\ttfamily\scriptsize,
    numberstyle=\tiny,
    stepnumber=1,
    numbersep=5pt,
    showstringspaces=false,
    breaklines=true,
    frame=single,
    literate=
     *{0}{{{\color{blue}0}}}{1}
      {1}{{{\color{blue}1}}}{1}
      {2}{{{\color{blue}2}}}{1}
      {3}{{{\color{blue}3}}}{1}
      {4}{{{\color{blue}4}}}{1}
      {5}{{{\color{blue}5}}}{1}
      {6}{{{\color{blue}6}}}{1}
      {7}{{{\color{blue}7}}}{1}
      {8}{{{\color{blue}8}}}{1}
      {9}{{{\color{blue}9}}}{1}
      {:}{{{\color{red}{:}}}}{1}
      {,}{{{\color{red}{,}}}}{1}
}

\lstdefinelanguage{xml}{
  basicstyle=\ttfamily\scriptsize,
  morestring=[b]",
  morecomment=[s]{<!--}{-->},
  morekeywords={xmlns,version,type, src, trg, id, topic_id}, 
  keywordstyle=\color{blue},
  stringstyle=\color{red},
  commentstyle=\color{gray},
  breaklines=true,
  frame=single
}

\title{AMELIA: A Family of Multi-task End-to-end Language Models for Argumentation
}

\author{
  Henri Savingy, Bruno Yun \\
  Universite Claude Bernard Lyon 1,\\
  CNRS, Ecole Centrale de Lyon, INSA Lyon, Université Lumiere Lyon 2,\\
LIRIS, UMR5205,\\
France\\
  \texttt{henri.savigny@etu.univ-lyon1.fr}, \texttt{\{bruno.yun\}@univ-lyon1.fr} \\
}

\begin{document}
\maketitle

\begin{abstract}
Argument mining is a subfield of argumentation that aims to automatically extract argumentative structures and their relations from natural language texts.
        This paper investigates how a single large language model can be leveraged to perform one or several argument mining tasks.
        
        Our contributions are two-fold.
        First, we construct a multi-task dataset by surveying and converting 19 well-known argument mining datasets from the literature into a unified format.
        Second, we explore various training strategies using Meta AI's Llama-3.1-8B-Instruct model: (1) fine-tuning on individual tasks, (2) fine-tuning jointly on multiple tasks, and (3) merging models fine-tuned separately on individual tasks.
        
        Our experiments show that task-specific fine-tuning significantly improves individual performance across all tasks.
        Moreover, multi-task fine-tuning maintains strong performance without degradation, suggesting effective transfer learning across related tasks.
        Finally, we demonstrate that model merging offers a viable compromise: it yields competitive performance while mitigating the computational costs associated with full multi-task fine-tuning.

\end{abstract}


\section{Introduction}

Argumentation theory aims to model, analyse, and automate argumentative reasoning by relying on formal representations of arguments and their relationships. In his seminal paper, \cite{dung1995acceptability} introduced the concept of \textit{abstract argumentation frameworks (AAFs)} where arguments are represented as abstract nodes in a graph and attacks between them are directed edges. He then defined various \textit{acceptability semantics} (such as preferred, complete, and grounded among others) to determine which subsets of arguments (known as \textit{extensions}) can be collectively accepted.
These acceptability semantics\footnote{We refer the reader to the guide by \cite{baroni2011introduction} for a comprehensive tutorial with guided explanations.}, offer specific criteria to select acceptable arguments and enable reasoning under uncertainty and inconsistency \citep{DBLP:conf/sum/CroitoruV13,DBLP:journals/argcom/YunVO22}. 

Building on Dung's framework, researchers have massively explored extensions of this abstract model with additional expressive components (such as support relations \citep{DBLP:conf/nmr/AmgoudCL04,DBLP:conf/ecsqaru/YunV21} or sets of attacking arguments \cite{DBLP:conf/comma/NielsenP06a,DBLP:conf/comma/YunVC20}) or the development of new semantics (e.g., to assess the strength of arguments in a more gradual manner beyond binary acceptability \citep{amgoud2022evaluation,DBLP:conf/aaai/YunVC20}).

These research efforts enhance the expressivity of the original framework and also contribute to more diverse applicability in real-world domains such as legal reasoning and political analysis \citep{amgoud2009using}.

However, a key limitation of approaches based on Dung’s AAFs lies in their reliance on the assumption that both arguments and their relations are explicitly specified. While manual formalisation may be feasible at small scale in expert-driven systems, the automatic extraction of arguments and their interrelations from large-scale, unstructured data remains an open research challenge.
    
        In recent years, argument mining (AM), the task of automatically identifying and extracting argumentative structures from natural language texts, has received growing attention in the natural language processing (NLP) community \citep{lawrence2020argument}. In more details, this task involves detecting argumentative components (such as claims and premises) and the relations between them. Given the pervasive role of argumentation in both written and spoken communication, the development of computational models capable of parsing and evaluating arguments has promising applications in downstream tasks such as decision-making, persuasion, fact-checking, and misinformation detection \citep{wang2025automated,figueras2024using}.
        
        Early approaches to argument mining relied on models such as structured support machines (SVMs) and recurrent neural networks (RNNs) \citep{niculae2017argument}. However, the development of deep-learning and transformer-based architectures, allowed argument mining systems to gradually improve their performance. For instance, \cite{mayer2020transformer} applied transformer-based models to mine arguments in healthcare-related texts, demonstrating that domain-adapted language models can capture argumentation patterns in specialised corpora. Moreover, with the recent development of large language models (LLMs), recent work has demonstrated the potential of prompting or fine-tuning LLMs specifically for argument mining \citep{gorur2024can,faugier2024assisted,cabessa2025argument,stahl2025arginstruct}. E.g., \cite{cabessa2025argument} showed that appropriately fine-tuned LLMs can outperform traditional architectures in extracting argumentative structure from natural language texts and \cite{stahl2025arginstruct} developed \textit{Arginstruct}, which fine-tuned LLMs using argumentation instruction tuning to enhance model performance in argumentative tasks.

        However, we argue that these previous approaches mainly focused on individual argument mining tasks (such as argument component or argument relation classification) limiting their applicability to more end-to-end or open-ended argumentative tasks. 
        Furthermore, while instruction tuning methods, such as the one used in \textit{Arginstruct} \citep{stahl2025arginstruct}, show promise for improving LLMs' performance in argumentative contexts, it largely depends on synthetic, LLM-generated instructions as training data, rather than leveraging real-world, human-annotated argumentation data. This may, in turn, limit the generalizability of the models' argumentative capabilities in real-world settings.

        Thus, this work aims to \textit{explore the potential of a single LLM to perform one or several argument mining tasks}.
        %
        In doing this, we explore two central research questions:
                \begin{itemize}
                   \item[\textbf{RQ1:}]                    
            ``\textit{To what extent does fine-tuning improve the performance of an LLM on argument mining tasks?}'' 
                    \item[\textbf{RQ2:}] 
             ``\textit{Can an LLM obtain good performance in multiple argument mining tasks? If yes, how?}''.
                \end{itemize}
        
        Our several contributions aimed at answering these two questions.
        Namely, we first surveyed and collected 19 real-world AM datasets from the literature and converted them into a standardised format suitable to train LLMs.
        Next, we fine-tuned and evaluated Meta AI's Llama-3.1-8B-Instruct LLM on eight identified argument mining tasks. 
        Lastly, to explore the capabilities of LLMs to perform multiple AM tasks, we explored (1) a model merging approach to combine the previously fine-tuned models and (2) fine-tuning jointly on multiple AM tasks.

Our experiments show that task-specific fine-tuning significantly improves individual performance across all tasks.
        Moreover, multi-task fine-tuning maintains strong performance without degradation, suggesting effective transfer learning across related tasks.
        Finally, we demonstrate that model merging offers a viable compromise: it yields competitive performance while mitigating the computational costs associated with full multi-task fine-tuning.

        This report is structured as follows. In Section \ref{sect:related_work}, we introduced the background and related work on large language models and argument mining. 
        In Section \ref{sect:dataset}, we describe the refinement of the existing dataset and the creation of our multi-task dataset. 
        In Section \ref{sec:llm-multitask}, we explain the training regiment for the several LLMs in the context of AM and their evaluations. 
        Finally, we conclude in Section \ref{sect:ccl}.

 \section{Related Work}
\label{sect:related_work}

In this section, we begin with a brief overview of prior work on large language models and their connection to argument mining. We then narrow our focus to recent studies that specifically apply LLMs to argument mining tasks. Finally, we present the various argument mining datasets available in the literature, which serve as the foundation for our experiments in the following sections.

\subsection{Large Language Models}

A language model is a computational system designed to generate or understand human language by estimating the probability of word sequences. This allows the model to predict, complete, or generate coherent text based on a given context.
            Early language models were purely statistical, relying on methods such as $n$-gram or skip-gram models \citep{mikolov2013efficient}. Over time, these approaches gave way to neural models, first using recurrent neural networks (RNNs) and later advancing to transformer-based architectures, which now form the backbone of modern large language models.
           Among transformer-based models, BERT (Bidirectional Encoder Representations from Transformers) stands out as a prominent encoder-only model. It introduced masked language modelling and bidirectional attention, achieving state-of-the-art performance on a wide range of natural language understanding tasks through fine-tuning.
            In 2019, GPT-2 \citep{radford2019language} sparked widespread public interest in generative language models. Built on a decoder-only architecture and trained using an autoregressive objective, GPT-2 demonstrated that scaling up both model size and training data could lead to strong performance across diverse tasks, without the need for task-specific supervision.

            Nowadays, we call \textit{large language models}, generative language models with many parameters (usually in billions) and trained on a vast amount of textual data to perform a wide range of natural language tasks. 
            The release of OpenAI's GPT-3 \citep{mann2020language} further demonstrated that increasing model scale improved zero-shot and few-shot performances. This led to the development of various open-source/closed-sourced models such as Meta AI's Llama models \citep{grattafiori2024llama}, Mistral-7B \citep{jiang2023mistral7b}, Alibaba's Qwen2.5\cite{yang2025qwen3}, and OpenAI's GPT-4 \citep{achiam2023gpt} among plenty of others which further improved performance and exhibited improved alignment and reasoning capabilities. 
            Further improvement can be seen in models such as Claude \citep{anthropic2025claudesonnet}, which integrated architectural changes to support extended thinking and Cogito \citep{deepCogito}, which explores human-inspired reasoning architecture. This race to better reasoning has been seen in other models such as DeepSeek-R1 \citep{guo2025deepseek} and Deepseek-V3 \citep{liu2024deepseek}, which emphasise reasoning through reinforcement learning training, and Alibaba's QwQ model \citep{yang2024qwq}, designed for code generation and theorem proving.

            As LLMs became more ubiquitous, controlling their performances for specific tasks became essential. 
            %
            On the one hand, research on prompt engineering, such as Chain-of-Thought (CoT) prompting \citep{wei2022chain}, which enhances reasoning by guiding an LLM to generate intermediate steps, has been shown to enhance accuracy on reasoning-intensive tasks. This was followed by more structured techniques like Graph-of-Thought (GoT) \citep{yao2023beyond} and Thread-of-Thought (ToT) \citep{zhou2023thread}, which aim to improve interpretability and robustness. 
            On the other hand, supervised Fine-Tuning (SFT) and instruction tuning have also emerged as powerful methods to improve LLM performances. \cite{chung2024scaling} have shown that scaling instruction tuning improves generalisation and task transfer, while PEFT (Performance Effective Fine Tuning) techniques such as LoRA \citep{hu2022lora} (and its many variants) achieve performance comparable to full fine-tuning with greatly reduced computational cost.

            
\subsection{Argument Mining \& Large Language Models}

 Argument mining is a field of computational argumentation that aims to automatically identify and structure argumentative discourse in natural language texts. The main argument mining tasks include the detection of argumentative components (e.g, claims or premises) or the classification of relations between them \citep{lawrence2020argument}.
Note that in this work , we also study argument quality assessment, which is one of the major related AM tasks. It consists in assessing the quality of claims and their revisions \citep{skitalinskaya2021learning}. Other variants are post quality \citep{wang2023contextual} and overall quality assessment \citep{wachsmuth2017computational}.

            Early work in AM relied on structured prediction methods and neural architecture, such as structured SVMs and recurrent neural networks for component identification and relation prediction \citep{niculae2017argument} and a relation-based approach to capture argumentative structure \citep{carstens2015towards}. 
            We also note that relation classification can also be improved via the use of argumentation scheme and formal logic \citep{jo2021classifying}. 
            More recent approaches leverage transformer-based architecture, which demonstrate strong performance across a range of argument mining tasks, with applications in domains such as healthcare \citep{mayer2020transformer}, political debates analysis \citep{menini2018never}, and fallacy detection \citep{helwe2023mafalda,goffredo2023argument}.

            End-to-end frameworks have also been explored. For example, \cite{lenz2020towards} introduces a system that automatically transforms natural language text into an argument graph while 
            \cite{morio2022end} and \cite{schulz2018multi} propose models trained via multi-task learning to perform argument mining in low-resource settings. 
            To help reduce the need for a large amount of data, cross-corpora and training in low-resource settings have also been studied, with methods using generalizability and transfer learning \citep{cocarascu2020dataset,wang2023contextual}.
        
            Generative approaches have also been in argumentation as a text-to-text generation task \citep{kawarada2024argument} or to perform structured extraction of complex argumentative structure, like argument quadruplet extraction \citep{guo2023aqe}. However, in this work, we restrict the scope of our investigation and purposely do not put our focus on generative tasks.

The use of LLMs in argument mining has shown progress across core tasks (such as argument component identification, relation classification, fallacy detection and quality assessment). Initial explorations have demonstrated the potential of LLM in capturing argumentative structures, while recent evaluations have assessed their performance across precise sub-tasks \citep{abkenar2024assessing,mirzakhmedova2024large,gorur2024can}.

            However, LLMs showed several limitations, such as difficulties in reliably detecting argumentative fallacies in natural settings \citep{ruiz2023detecting} or logical inconsistency in generated fallacy annotation \citep{yeh2024cocolofa}.
            As a solution, fine-tuning and in-context learning have been explored to adapt LLMs to argument mining. \cite{cabessa2024context,cabessa2025argument} showed that fine-tuned models improve in the task of argument component classification and relation classification. Intruction-tuned variant such as ArgInstruct \cite{stahl2025arginstruct}, further enhances LLM capabilities by aligning them more closely with the reasoning requirement of argumentative tasks. However, they only use synthetic LLM-generated instructions to perform their instruction tuning.
            We argue that this may limit the generalizability of the models' argumentative capabilities in real-world settings. In our approach, we make use of real datasets from the literature which we introduce in the next section.

\subsection{Survey of Datasets in Argument Mining}
\label{sec:surveydataset}

In the following, we present the 19 datasets collected in the literature, which we will convert and use for fine-tuning LLMs. They are presented in alphabetical order, with illustrative examples provided for only a subset for the sake of brevity.

                \begin{figure*}[!h]
                \centering
                \includegraphics[width=0.8\linewidth]{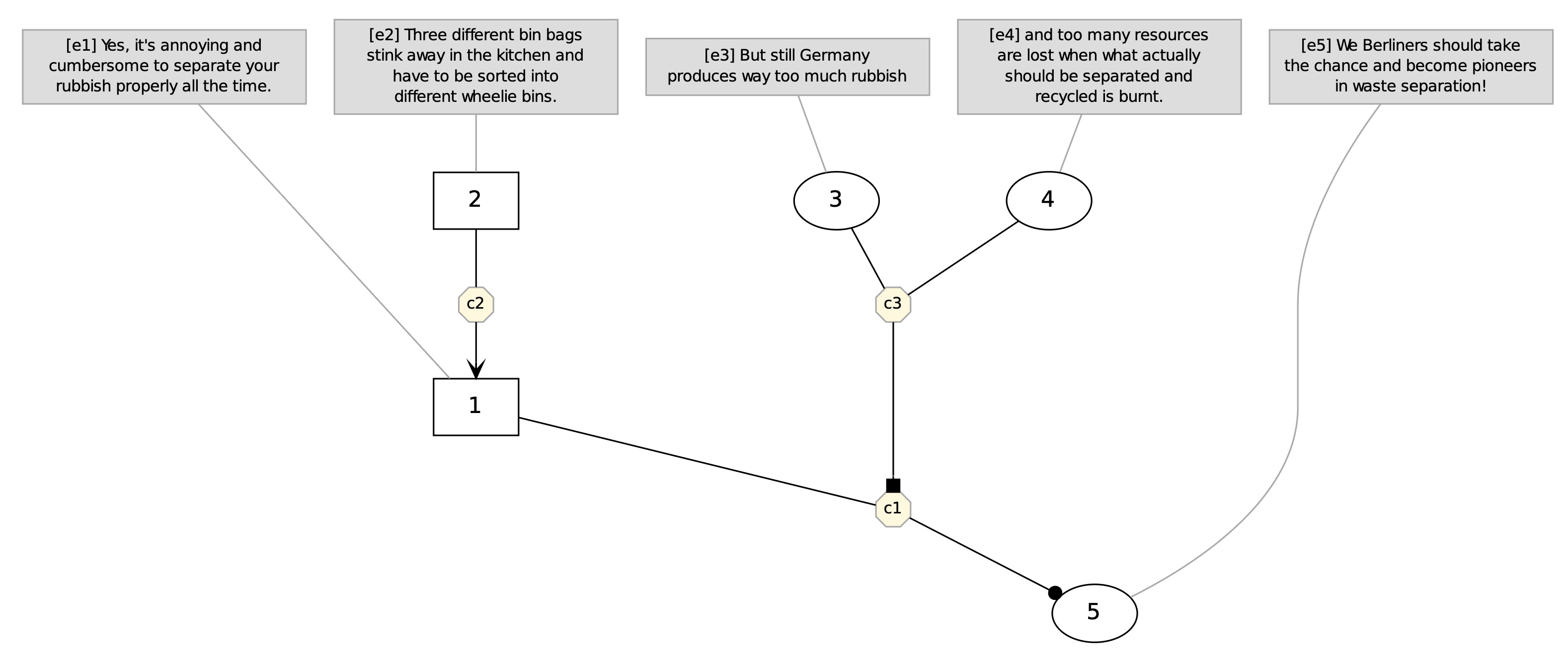}
                \caption{Representation of the micro-level argument graph on the topic of waste separation from Microtext part 1.}
                \label{fig:entrymicrotext1}
            \end{figure*}

   \paragraph{AbstRCT.} Proposed by \cite{mayer2020transformer}, this dataset targets argument mining in healthcare. It includes 500 medical texts annotated for argumentative component (major claim, claim and premises) and argument relations (attack or support). The dataset highlights challenges like domain-specific vocabulary, evidence scarcity and interpretability.

            \paragraph{AQM.} \cite{guo2023aqe} presented AQM as a dataset for their Argument Quadruplet Extraction (AQE) task, which involves identifying four elements in a statement: the topic, stance, opinion and rationale. The data consist of 34,369 sentences from 801 articles annotated for three argument components: claim, evidence and stance.

            \paragraph{ArgSum.} \cite{li2024side} introduced ArgSum as a comprehensive multi-task dataset designed for end-to-end argument summarization and evaluation. It contains user-generated discussions with associated stances, generated summaries, and human or model-generated quality judgments. The dataset covers multiple tasks such as argument component extraction, stance detection, summarisation and summary evaluation. Its structure supports joint training across those tasks, enabling multi-task learning.

            \paragraph{ComArg.} \cite{boltuvzic2014back} developed this dataset by extracting data from online debate forums and social platforms. It includes 2,436 comments annotated for stance and argument recognition. This dataset focuses on short, user-generated responses to controversial questions. This dataset supports tasks such as argument relation classification and stance detection. 

            \paragraph{CoCoLoFa.} \cite{yeh2024cocolofa} introduced CoCoLoFa as a dataset containing 7,706 news comments from 648 news articles annotated for eight common logical fallacies (each comment is annotated with one fallacy among the eight selected fallacies), which were verified with the help of an LLM and human annotators. It covers various fallacy types and is situated in real-world opinionated discussions, such as reader comments on a news article.

            \paragraph{Dagstuhl-15512 ArgQuality.} \cite{wachsmuth2017computational} proposed this dataset to assess argument quality in natural language across several dimensions such as clarity, cogency, sufficiency and effectiveness. The dataset consists of 320 arguments rated by human annotators over the different quality dimensions. This dataset helps quantify what makes an argument ``good'' and ``persuasive''.

            \paragraph{FEVER.} \cite{thorne2018fever} compiled a dataset consisting of 185,445 claims for fact verification by pairing claims with evidence from Wikipedia. Though primarily designed for factual verification, the dataset's structure overlaps with tasks in argument mining. Each claim is annotated as ``supported'', ``refuted'' or ``not enough information'' based on the retrieved evidence.

            \paragraph{IAM.} \cite{cheng2022iam} presented IAM, a large-scale dataset designed for multiple tasks of argument mining, including argument component identification, relation classification and stance detection. It includes 69,666 sentences spanning multiple domains and having extensive annotations. IAM supports both individual and multi-task learning. It is one of the most comprehensive datasets currently available for training and benchmarking argument mining models.

            \paragraph{IBM Claim-polarity.} \cite{bar2017stance} proposed this dataset of 2,394 claims associated with 55 topics annotated for stance classification. The claims are labelled for polarity (support, oppose, neutral) and contextual dependency.
            
            \paragraph{IBM Type.} \cite{aharoni2014benchmark} introduced this dataset with claim and evidence annotations across 33 controversial topics, including 2,883 arguments across 586 documents from Wikipedia. The focus is on distinguishing argument components and associating claims with relevant evidence.

            \paragraph{IBM Claim.} \cite{levy2018towards} constructed this dataset, consisting of 2500 claims, by extracting data from web sources associated with 50 distinct topics. It focused on claim detection and aims to bridge the gap between large, noisy web data and structured argumentative search.
            
            \paragraph{IBM Evidence.} \cite{shnarch2018will} presented this dataset of 5,785 sentences annotated for evidence (either supporting or contesting the topic) across 83 topics.
            
            \paragraph{IBM Argument.} \cite{shnarch2020unsupervised} proposed this dataset to address the challenge of domain adaptation in argument mining. The datasets enable models to generalise to unfamiliar domains while retaining interpretability. The dataset includes 700 sentences annotated as to whether they contain an argument for the given topic. The sentence annotations span across 20 topics, and the rules provide insight into how argument patterns across multiple domains.
            
            \paragraph{MAFALDA.} \cite{helwe2023mafalda} released MAFALDA as a benchmark dataset for logical fallacy detection and classification, covering over 30 fallacy types. It contains a mix of real and synthetic texts, annotated by experts and LLM-assisted crowd-workers. It includes 200 texts in which sentences have been with one or more fallacies. We give a single entry of this dataset in Example \ref{ex:mafalda}.
            
                \begin{mdframed}[backgroundcolor=lightgray!40, roundcorner=10pt]
                    \begin{example} \label{ex:mafalda}
                        MAFALDA entry identifying two fallacies (false dilemma and hasty generalization) from a post claiming that because a bar in Thurles wasn’t attacked over an ad showing Jesus with a pint, Christians are not as sensitive as Muslims.
                        
                        \begin{lstlisting}[language=json]
{
    "text": "TITLE: Bar in Thurles in trouble over ad featuring Jesus with a pint. Christians are slowly becoming bigger snowflakes than Muslims. POST: So was the bar burned by a mob and the owner killed? If not, Christians have a ways to go before they are on par with Muslims.\n",
    "labels": [[139, 265, "false dilemma"], [139, 265, "hasty generalization"]],
    "comments": ["Hasty Generalization: S=Christian/Muslim extremists, P=All Christians/Muslims", "false dilemma: X = Christians burned the bar and are like Muslims, Y = Christians did'nt burn the bar and are not like Muslims"], 
    "sentences_with_labels": "{\"TITLE: Bar in Thurles in trouble over ad featuring Jesus with a pint.\": [[\"nothing\"]],\"Christians are slowly becoming bigger snowflakes than Muslims.\": [[\"nothing\"]], \"POST:\": [[\"nothing\"]], \"So was the bar burned by a mob and the owner killed?\": [[\"hasty generalization\", \"false dilemma\"]], \"If not, Christians have a ways to go before they are on par with Muslims.\": [[\"hasty generalization\", \"false dilemma\"]]}"
}
                        \end{lstlisting}

                    \end{example}
                \end{mdframed}

                \paragraph{Microtext part 1} \cite{peldszus2015annotated} released a structured dataset composed of 112 microtexts, each containing a full argument annotated (claim, premises) and their relations (attack, support). We give a single entry of this dataset in Example \ref{ex-microtextpart1}.

                \begin{mdframed}[backgroundcolor=lightgray!40, roundcorner=10pt]
                     \begin{example}\label{ex-microtextpart1}
                        Microtext part 1 entry about the topic of waste separation. Five elementary discourse units (edu) are identified and associated with argumentative discourse units (adu). Relations between argumentative discourse units and their relations are specified in $c_1, c_2, c_3$ and $c_4$. The corresponding micro-level argument graph is represented in Figure \ref{fig:entrymicrotext1}.

\begin{lstlisting}[language=xml]
<?xml version='1.0' encoding='UTF-8'?>
<arggraph id="micro_b001" topic_id="waste_separation" stance="pro">
  <edu id="e1"><![CDATA[Yes, it's annoying and cumbersome to separate your rubbish properly all the time.]]></edu>
  <edu id="e2"><![CDATA[Three different bin bags stink away in the kitchen and have to be sorted into different wheelie bins.]]></edu>
  <edu id="e3"><![CDATA[But still Germany produces way too much rubbish]]></edu>
  <edu id="e4"><![CDATA[and too many resources are lost when what actually should be separated and recycled is burnt.]]></edu>
  <edu id="e5"><![CDATA[We Berliners should take the chance and become pioneers in waste separation!]]></edu>
  <adu id="a1" type="opp"/>
  <adu id="a2" type="opp"/>
  <adu id="a3" type="pro"/>
  <adu id="a4" type="pro"/>
  <adu id="a5" type="pro"/>
  <edge id="c6" src="e1" trg="a1" type="seg"/>
  <edge id="c7" src="e2" trg="a2" type="seg"/>
  <edge id="c8" src="e3" trg="a3" type="seg"/>
  <edge id="c9" src="e4" trg="a4" type="seg"/>
  <edge id="c10" src="e5" trg="a5" type="seg"/>
  <edge id="c1" src="a1" trg="a5" type="reb"/>
  <edge id="c2" src="a2" trg="a1" type="sup"/>
  <edge id="c3" src="a3" trg="c1" type="und"/>
  <edge id="c4" src="a4" trg="c3" type="add"/>
</arggraph>
                         \end{lstlisting}

                     \end{example}
                \end{mdframed}

                \paragraph{Microtext part 2} \cite{skeppstedt2018more} extended Microtext part 1 with 171 texts by crowd-sourcing new argumentative texts under controlled prompts. The goal was to enlarge the original dataset while maintaining its clarity and consistency in argument structure.

            \paragraph{Nixon-Kennedy Debates.} \cite{menini2018never} curated and annotated the 1960 US Nixon-Kennedy presidential debates with a focus on argumentation strategies in political speech. The dataset includes 1,907 argument pairs covering 5 topics annotated for argumentative relations and rhetorical patterns, enabling analysis of persuasive techniques and discourse dynamics.

             \paragraph{Node.} \cite{cabrio2014node} introduced the Node dataset, comprising 260 arguments extracted from online and encyclopedic sources. Each argument is annotated with acceptability judgments based on its coherence and logical structure.
            
            \paragraph{Persuasive Essays.} \cite{stab2017parsing} provided a dataset containing persuasive essays annotated with argument components (major claim, claim, premise) and their relations (support, attack). It includes 402 essays written by students and annotated by three annotators (two non-expert and one expert annotator), offering a consistent structure and real-world argumentative writing.

   \begin{mdframed}[backgroundcolor=lightgray!40,  roundcorner=10pt]
                \begin{example}\label{ex:json}
                    Corresponding \texttt{Json} entry for the Microtext part 1 entry described in Example \ref{ex-microtextpart1}.

                    \begin{minipage}[t]{0.95\textwidth}

\begin{lstlisting}[language=json]
    {
        "id": "micro_b001",
        "topic": "waste_separation",
        "stance": "pro",
        "edu": [
            {"e1": "Yes, it's annoying and cumbersome to separate your rubbish properly all the time."},
            {"e2": "Three different bin bags stink away in the kitchen and have to be sorted into different wheelie bins."},
            {"e3": "But still Germany produces way too much rubbish"},
            {"e4": "and too many resources are lost when what actually should be separated and recycled is burnt."},
            {"e5": "We Berliners should take the chance and become pioneers in waste separation!"}
        ],
        "adu": [
            {"e1": "a1", "stance": "opp", "label": "Premise"},
            {"e2": "a2", "stance": "opp", "label": "Premise"},
            {"e3": "a3", "stance": "pro", "label": "Premise"},
            {"e4": "a4", "stance": "pro", "label": "Premise"},
            {"e5": "a5", "stance": "pro", "label": "Claim"}
        ], 
        "relations": [
            {"id": "c1", "src": "a1", "trg": "a5", "type": "reb"}, 
            {"id": "c2", "src": "a2", "trg": "a1", "type": "sup"},
            {"id": "c3", "src": "a3", "trg": "c1", "type": "und"},
            {"id": "c4", "src": "a4", "trg": "c3", "type": "add"}
        ]
    }
\end{lstlisting}
                    \end{minipage}
                \end{example}
            \end{mdframed}

\section{A New Dataset for Multi-Task Argument Mining}
\label{sect:dataset}

The 19 datasets identified in Section \ref{sec:surveydataset} cannot be directly used for training as they possess widely different specificities and formats. In this section, we explain our approach to unify (Section \ref{sec:jsonl}) and exploit these datasets for eight argument mining tasks we consider (Section \ref{subsect:task_and_datasets}). 

\subsection{Unifying Existing AM Datasets} \label{sec:jsonl}
                    
                    To facilitate the fine-tuning and testing of LLMs (as well as reproducibility), we first unify each of the datasets under a handcrafted \texttt{jsonl} format. This unique format eliminates the need for multiple format-specific parsers. 
                    Moreover, data were filtered (removing unnecessary tags, etc.), and important informational keywords were added.
                    Example \ref{ex:json} shows how the entry from Example \ref{ex-microtextpart1} is converted into a \texttt{json} entry. 
                    
                    We encourage the reader to explore the different converted datasets in our GitHub repository: \url{https://github.com/brunoyun/amelia}.

\subsection{Argument Mining Tasks Considered} \label{subsect:task_and_datasets}

\begin{figure*}[!h]
    \begin{center}
        \includegraphics[width=0.6\textwidth]{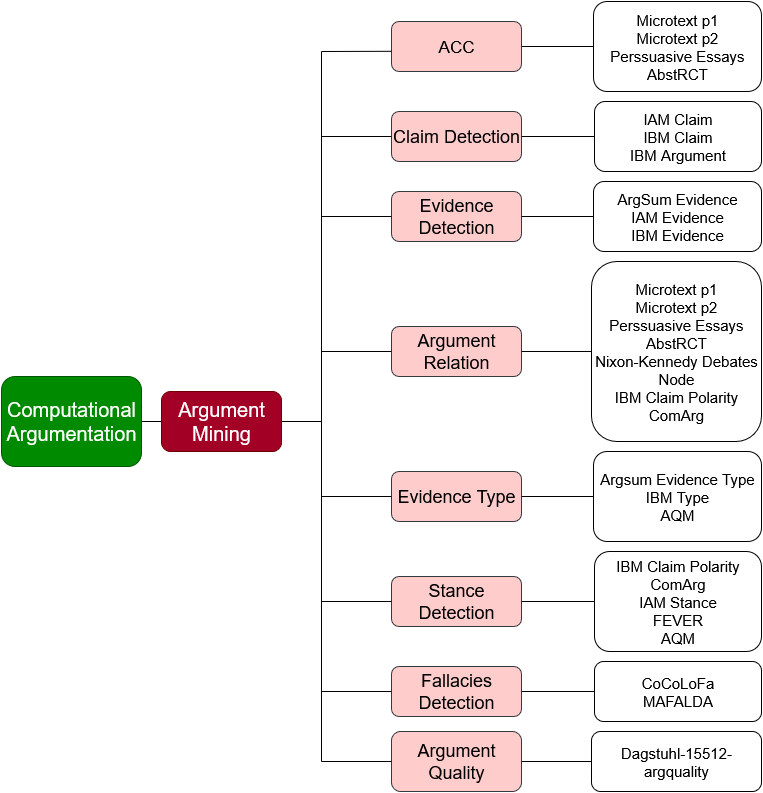}
        \caption{Graph of the different argumentation tasks and the associated datasets.}
        \label{fig:Diag_arg_task}
    \end{center}
    
\end{figure*}

All the studied tasks (in light red boxes) and associated datasets (in white boxes) are represented in Figure \ref{fig:Diag_arg_task}. 
            We selected those AM tasks by surveying the literature on AM and restricting ourselves to tasks which can be converted into classification tasks.
            Note that some datasets (e.g., AQM) are reused for different tasks in different ways.

         In the rest of this section, we will describe and formalise each of those tasks.

            \paragraph{Argument Component classification (ACC).}
                Argumentative discourse units represent the smallest components within a text that contribute to its argumentative structure.
                ACC is the task of classifying argument component as either  ``premises'' or ``claim''. This classification task does not address the distinction between argument and non-argumentative materials.
                We formalised this task as:
                \begin{center}
                    \begin{minipage}{0.9\linewidth}
                        \underline{\textbf{Input}}: A topic $t$ and a sentence $s$ \\
                        \underline{\textbf{Output}} : $o\in \{\mathtt{Claim}, \mathtt{Premises}\}$
                    \end{minipage}
                \end{center}
                
                We fine-tuned and evaluated this task on datasets including Microtext part 1 \& 2, Persuasive Essays, and AbstRCT. Example \ref{ex:ACC} shows an example of input/output for the ACC task from Microtext part 1.
                
                \begin{mdframed}[backgroundcolor=lightgray!40, roundcorner=10pt]
                    \begin{example}\label{ex:ACC}
                        Consider this example from Microtext part 1.

                        \noindent\underline{\textbf{Input}} : \begin{itemize}
                            \item $t$: ``introduce capital punishment''
                            \item $s$: ``If Germany were to introduce the death penalty, convicted felons could not be paroled or break out of prison and commit new felonies''
                        \end{itemize}
                        \underline{\textbf{Output}} : $\mathtt{Premises}$

                        \medskip
                        
                        \noindent
                        The sentence $s$ provides a reason supporting the topic $t$, rather than asserting the main point or an opinion about the topic, thus leading to the classification as ``Premises''.
                    \end{example}
                \end{mdframed}

            \paragraph{Claim detection (CD).}
                A claim is a statement that asserts something to be true or false. In argument mining, a claim serves as a central component that forms the basis of reasoning and debate. The CD task aims to identify and extract claims relevant to a given debate's topic from texts.
                We formalised this task as:
                
                \begin{center}
                    \begin{minipage}{0.9\linewidth}
                        \underline{\textbf{Input}} : A topic $t$ and sentence $s$\\
                    \underline{\textbf{Output}} : $ o \in \{\mathtt{Claim}, \mathtt{Non}$-$\mathtt{Claim}\}$
                    \end{minipage}
                \end{center}
                
                We fine-tuned and evaluated the claim detection task on datasets including IAM Claim, IBM Claim, and IBM Argument.
                Example \ref{ex:CD} shows an example of input/output for the CD task from IAM Claim.
                
                \begin{mdframed}[backgroundcolor=lightgray!40, roundcorner=10pt]
                    \begin{example}\label{ex:CD}
                        Consider this example from IAM Claim.

                        \noindent\underline{\textbf{Input}} : \begin{itemize}
                            \item $t$: ``Shouldn't the sale of violent video games be banned''
                            \item $s$: ``It has nothing to do with video games or Paxil, and my son's no murderer.''
                        \end{itemize}
                        \underline{\textbf{Output}} : $\mathtt{Non}$-$\mathtt{Claim}$

                        \medskip

                        \noindent Despite being an opinionated sentence, $s$ does not express a claim about the topic.
                    \end{example}
                \end{mdframed}

            \paragraph{Evidence detection (ED).}
                An evidence refers to any information or data that either supports or challenges a claim. In argument mining, the task of evidence detection focuses on the identification and extraction of relevant pieces of text that help validate or refute claims.
                We formalised ED as:
                
                \begin{center}
                    \begin{minipage}{0.9\linewidth}
                        \underline{\textbf{Input}} : A topic $t$, a claim $c$, and a sentence $s$\\
                        \underline{\textbf{Output}} : $o\in \{\mathtt{Evidence}, \mathtt{Non}$-$\mathtt{evidence}\}$
                    \end{minipage}
                \end{center}
                
                We fine-tuned and evaluated the evidence detection task on the ArgSum Evidence, IAM Evidence, and IBM Evidence datasets.
                Example \ref{ex:ED} shows an example of input/output for the ED task from ArgSum Evidence.
                
                \begin{mdframed}[backgroundcolor=lightgray!40, roundcorner=10pt]
                    \begin{example}\label{ex:ED}
                        Consider this example from ArgSum Evidence.

                        \noindent\underline{\textbf{Input}} : \begin{itemize}
                            \item $t$: ``We should ban private military companies''
                            \item $c$: ``Private military companies' main interest is profit''
                            \item $s$: ``A report by Human Rights Watch in 2006 found that private military companies often prioritise profit over other considerations, including ethical and legal considerations.''
                        \end{itemize}
                        \underline{\textbf{Output}} : $\mathtt{Evidence}$

                        \medskip

                        \noindent Here, $s$ cite a report that validate the claim $c$.
                    \end{example}
                \end{mdframed}
                
            \paragraph{Argument relation classification (AR).}
                The objective of argument relation classification is to determine whether a given pair of arguments is connected through an argumentative relationship. Given a pair of arguments (a source argument and a target argument), the task is to classify the relationship from the source argument to the target argument as either ``attack'', ``support'' or ``no relation''.
                Formally, this task is described as :
                \begin{center}
                    \begin{minipage}{0.9\linewidth}
                        \underline{\textbf{Input}} : A topic $t$, a source argument $a_{src}$, and a target argument $a_{trg}$\\
                        \underline{\textbf{Output}} : $o \in \{\mathtt{attack}, \mathtt{support}, \mathtt{no\;relation}\}$
                    \end{minipage}
                \end{center}
                We fine-tuned and evaluated the argument relation classification task on datasets including Microtext part 1 \& 2, Persuasive Essays, AbstRCT, Nixon-Kennedy Debates, Node, IBM Claim-polarity and ComArg.
                Example \ref{ex:AR} shows an example of input/output for the AR task from Node.

                \begin{mdframed}[backgroundcolor=lightgray!40, roundcorner=10pt]
                    \begin{example}\label{ex:AR}
                        Consider this example from Node.

                        \noindent\underline{\textbf{Input}} : \begin{itemize}
                            \item $t$ : ``Tablet''
                            \item $a_{src}$: ``Tablets help students learn more material faster than textbooks. Technology-based instruction can reduce the time students take to reach a learning objective by 30-80\%, according to the US Department of Education and studies by the National Training and Simulation Association.''
                            \item $a_{trg}$: ``People who read print text comprehend more, remember more, and learn more than those who read digital text.''
                        \end{itemize}
                        \underline{\textbf{Output}} : $\mathtt{attack}$

                        \medskip

                        \noindent $a_{src}$ disagrees with $a_{trg}$ and uses a study to attack $a_{trg}$.
                    \end{example}
                \end{mdframed}

            \paragraph{Evidence type classification (ET).}
                The evidence types refer to the different categories of evidence that can either support or challenge a claim. Common types of evidence include anecdotal, expert opinion, explanation and study.
                Formally, this task is described as :
                
                \begin{center}
                    \begin{minipage}{0.9\linewidth}
                        \underline{\textbf{Input}}: A topic $t$, a claim $c$, and an evidence $e$\\
                        \underline{\textbf{Output}}: $o\in \{\mathtt{NONE},\mathtt{ANECDOTAL}, \mathtt{EXPERT},$ $\mathtt{EXPLANATION},$ $\mathtt{STUDY}\}$
                    \end{minipage}
                \end{center}
                
                This task was fine-tuned and evaluated on datasets including ArgSum Evidence Type, IBM Type, and AQM.
                Example \ref{ex:ET} shows an example of input/output for the ET task from AQM.
                
                \begin{mdframed}[backgroundcolor=lightgray!40, roundcorner=10pt]
                    \begin{example}\label{ex:ET}
                        Consider this example from AQM.

                        \noindent\underline{\textbf{Input}} : \begin{itemize}
                            \item $t$: ``Shouldn't the sale of violent video games be banned''
                            \item $c$: ``Conversely, playing violent video games had significantly more hurtful behaviors in children than the children who played prosocial games.''
                            \item $e$: ``Results indicated that playing prosocial games significantly more helpful behaviors in children than those who played violent video games.''
                        \end{itemize}
                        \underline{\textbf{Ouput}} : $\mathtt{STUDY}$

                        \medskip

                        \noindent Here, the evidence $e$ support the claim $c$ using the result from a study.
                    \end{example}
                \end{mdframed}
                
            \paragraph{Stance detection (SD).}
                A stance reflects a point of view on a debated subject, expressed as either support or opposed. The task of stance detection is to determine whether an argument supports or opposes a specific topic.
                Formally, this task is described as :
                \begin{center}
                    \begin{minipage}{0.9\linewidth}
                        \underline{\textbf{Input}} : A topic $t$ and a sentence $s$\\
                        \underline{\textbf{Output}} : $o \in \{\mathtt{For}, \mathtt{Against}\}$
                    \end{minipage}
                \end{center}
                
                We fine-tuned and evaluated the stance detection task using datasets such as IBM Claim-Polarity, ComARg, IAM Stance, FEVER, and AQM.
                Example \ref{ex:SD} shows an example of input/output for the SD task from IAM Stance.
                
                \begin{mdframed}[backgroundcolor=lightgray!40, roundcorner=10pt]
                    \begin{example}\label{ex:SD}
                        Consider this example from IAM Stance.

                        \noindent\underline{\textbf{Input}} : \begin{itemize}
                            \item $t$: ``Should all museums be opened for free'' 
                            \item $s$: ``Free access is essential to provide freedom of cultural and educational opportunity.''
                        \end{itemize}
                        \underline{\textbf{Output}} : $\mathtt{For}$

                        \medskip

                        \noindent The sentences $s$ agree with the topic, thus leading to the classification $\mathtt{For}$.
                    \end{example}
                \end{mdframed}
            \paragraph{Fallacies detection (FD).}
                A fallacy is an argument where the premises do not entail the conclusion. The goal of fallacy detection is to identify whether a given argument contains a fallacy or not. In the latter case, the output is $\mathtt{none}$. If the argument is identified as fallacious, the task further involves classifying the kind of fallacy it contains.
                Formally, this task is described as:
                
                \begin{center}
                    \begin{minipage}{0.9\linewidth}
                        \underline{\textbf{Input}}: A sentence $s$\\
                        \underline{\textbf{Output}}: $o\in \{\mathtt{none}, \mathtt{appeal\:to\:fear},$ $\mathtt{hasty\:generalization},$ $ \mathtt{appeal\:to\:worse\:problem},$ $\mathtt{appeal\:to\:authority},$ $\mathtt{false\:causality},$ $\mathtt{appeal\:to\:tradition},$ $\mathtt{ad\:populum},$ $\mathtt{guilt\:by\:association},$ $\mathtt{causal\:oversimplification},$ $\mathtt{false\:dillema},$ $\mathtt{appeal\:to\:ridicule},$ $\mathtt{false\:analogy},$ $\mathtt{slippery\:slope},$ $\mathtt{appeal\:to\:majority},$ $\mathtt{appeal\:to\:nature},$ $\mathtt{straw\:man},$ $\mathtt{circular\:reasoning},$ $\mathtt{equivocation},$ $\mathtt{ad\:hominem}\}$
                    \end{minipage}
                \end{center}
                
                We fine-tuned and evaluated the fallacy detection task on two datasets CoCoLoFa and MAFALDA.
                Example \ref{ex:FD} shows an example of input/output for the FD task from CoCoLoFa.
                Since MAFALDA includes instances with multiple ground-truth fallacies, we define two sub-tasks, denoted as $FD_{Single}$ and $FD_{Multi}$. The first considers cases with a single ground-truth fallacy, while the second accounts for cases with multiple fallacies. In $FD_{Multi}$, a prediction is considered correct if it belongs to the set of ground-truth fallacies. Recall is then computed as the proportion of ground-truth fallacies correctly identified across multiple LLM predictions.

                \begin{mdframed}[backgroundcolor=lightgray!40, roundcorner=10pt]
                    \begin{example}\label{ex:FD}
                        Consider this example from CoCoLoFa.

                        \noindent\underline{\textbf{Input}} : \begin{itemize}
                            \item $s$: ``People must come together and take this threat seriously, like this honourable man. Denialism needs to be completely shut down. All the experts agree, climate change is real and is a massive threat to the whole planet. As Obama says, no challenge poses a greater threat to future generations than climate change.''
                        \end{itemize}
                        \underline{\textbf{Output}} : $\mathtt{appeal\:to\:authority}$

                        \medskip

                        The sentence $s$ relies on the opinion of an authority figure who may not have relevant expertise, thus leading to the classification as $\mathtt{appeal\:to\:authority}$.
                    \end{example}
                \end{mdframed}
          
            \paragraph{Argument quality assessment (AQ).}
                Argument quality refers to how good an argument is; it indicates the degree to which an argument is considered strong and effective. This quality is evaluated based on 15 different quality dimensions: \textit{overall quality, local acceptability, appropriateness, arrangement, clarity, cogency, effectiveness, global acceptability, global relevance, global sufficiency, reasonableness, local relevance, credibility, emotional appeal}, and \textit{sufficiency}.
                The goal of argument quality assessment is to evaluate an argument by rating it as low, average or good across each of the 15 quality dimensions.
                Formally, this task is described as :
                \begin{center}
                    \begin{minipage}{0.9\linewidth}
                        \underline{\textbf{Input}}: A topic $t$, a stance $s$, an argument $a$, and a quality dimension $q$ with its textual definition $d_{q}$\\
                        \underline{\textbf{Output}}: $o\in \{\mathtt{Low}, \mathtt{Average}, \mathtt{High}\}$
                    \end{minipage}
                \end{center}
                
                We fine-tuned and evaluated this task using the Dagstuhl-15512 ArgQuality dataset.
                Example \ref{ex:AQ} shows an example of input/output for the AQ task from Dagstuhl-15512 ArgQuality Corpus.
                
                \begin{mdframed}[backgroundcolor=lightgray!40, roundcorner=10pt]
                    \begin{example}\label{ex:AQ}
                        Consider this example from Dagstuhl-15512 ArgQuality Corpus.

                        \noindent\underline{\textbf{Input}} : \begin{itemize}
                            \item $t$: ``personal-pursuit-or-advancing-the-common-good''
                            \item $s$: ``advancing-the-common-good''
                            \item $a$: ``I feel like advancing the common good is better than personal pursuit because most people will look out for themselves rather than look out for others. Looking out for yourself before looking out for others won't do a lot of good in the long run. I believe in the what goes around comes around thing, so when you do something for other people before you try to please yourself, not only does it help them, but something good will also come back to you, helping you better yourself.''
                            \item $q$: ``clarity''
                            \item $d_q$: ``Argumentation has a clear style if it uses correct and widely unambiguous language as well as if it avoids unnecessary complexity and deviation from the issue''
                        \end{itemize}
                        \underline{\textbf{Output}} : $\mathtt{Average}$

                        \medskip

                        \noindent Here, $s$ use ambiguous language and add complexity to the argument by contradicting itself in the first sentences, hence the $\mathtt{Average}$ clarity.
                    \end{example}
                \end{mdframed}

\subsection{Creating AM Task-Specific Datasets}

We created task-specific dataset for each of the eight tasks identified in Section \ref{subsect:task_and_datasets} by extracting the corresponding input/output from several datasets (as shown in Figure \ref{fig:Diag_arg_task}) using a specific methodology. 

First, we divided each of the 19 datasets into three splits: training ($60\%$), validation ($20\%$) and test split ($20\%$). 

Then, we perform a sampling on each of those 3 split to obtain the data needed to fine-tuned, validate, and test the model for our eight tasks.
More precisely, the sampling was performed on each corresponding split of each dataset corresponding to a specific task. Given the disparities in dataset size and class distribution, the sampling procedure was designed to ensure an equal number of instances per class within each task, mitigating class imbalance. Additionally, the sampling preserved the original proportion of each dataset split, maintaining the original proportion of examples contributed by each dataset.\\
                More formally, let us consider a task $t\in \{ADUC, CD, ED,$ $ARC,$ $ET,$ $SD, FD, AQ\}$ with the corresponding set of datasets $D_t$, e.g., $D_{FD} = \{$CoCoLoFa, MAFALDA$\}$.
                For a dataset $d \in D_t$, we use the notation $Class(d)$ for the set of class labels in the dataset. We write $CL_{D_t} = \bigcup_{d \in D_t} Class(d)$ for the set of all class labels within $D_t$.
                Moreover, for any dataset $d \in D_t$ and class $c \in Class(d)$, $NumEl(c, d)$ is the number of instances of class $c$ in $d$.

                To retain the orignal dataset contributions, we compute for each dataset $d\in D_t$ the sampling ratio $r_{c,d,t}$ for a class label $c \in CL_{D_t}$: $$ r_{c,d,t} = \frac{NumEl(c, d)}{\sum_{d'\in D_t} NumEl(c, d')}$$
                Subsequently, the number of instances of class $c$ to sample from $d$ is computed as: $$sample_{c,d,t} = r_{c,d,t} \times \frac{n_{sample}}{|Cl_{D_t}|}$$ with $n_{sample}$ the number of sample needed.

                This sampling method was used to generate $4000$ training samples and $800$ samples for both validation and test sets.

                Table \ref{table:ex_sampling_cd} shows an example of sampling the IAM Claim (69,666 elements), IBM Claim (2500 elements), and IBM Arguments (700 elements) datasets used for the Claim Detection (CD) task. We can notice that the train, validation, and test datasets are balanced but keep the same proportion of elements from each dataset.
                
                \begin{table*}[ht]
                    \begin{center}
                        \resizebox{\textwidth}{!}{
                            \begin{tabular}{| c |c | c | c | c || c | c | c | c |}
                                \hline
                                Split     &\multicolumn{4}{c||}{Train}                     & \multicolumn{4}{c|}{Validation \& Test} \\
                                \hline
                                Datasets  & IAM Claim & IBM Claim & IBM Arguments & Total  & IAM Claim & IBM Claim & IBM Arguments & Total \\
                                \hline
                                Claim     & $1659$    & $252$     & $89$          & $2000$ & $333$     & $50$      & $18$          & $401$ \\
                                Non-Claim & $1933$    & $54$      & $13$          & $2000$ & $387$     & $11$      & $3$           & $401$ \\
                                \hline
                                Total     & $3592$    & $306$     & $102$         & $4000$ & $720$     & $61$      & $21$          & $802$ \\
                                \hline
                            \end{tabular}
                        }
                        \caption{Sampling example for the task of Claim Detection. The Validation and Test split have the same number of instances}
                        \label{table:ex_sampling_cd}
                    \end{center}
                \end{table*}
\section{Large Language Models for (Multi-Task) Argument Mining}
\label{sec:llm-multitask}

In this section, we describe the experiments to answer our two research questions and our results.

\subsection{Experiment setup}
\label{sec:exp-setting}
For all of our experiments, we further fine-tuned Llama3.1-8B-Instruct, an already fine-tuned version of the base Meta AI's Llama 3.1-8B for instruction following. 
Llama 3.1-8B-Instruct was selected for all experiments because it strikes a strong balance between performance, efficiency, and accessibility. The open-source model is large enough to capture complex reasoning and language patterns, making it suitable for tasks that require nuanced understanding, while still being lightweight enough to run reliably with reasonable computational resources. Its instruction-tuned design ensures consistent and helpful responses across diverse prompts, which was essential for maintaining reproducibility and comparability across experiments. Using a single, well-established model throughout also guarantees methodological consistency and avoids confounding effects from model variation.

Moreover, a uniform prompt format for each task was used for both fine-tuning and inference; this format consists of a task description and an explicit specification of the expected output format. 
We used the special token \texttt{<|ANSWER|>} to delimitate the LLM's answer. This choice was inspired by the syntax of special tokens used in the Llama3.1 model's prompt template.
The first part of the prompt describes the objectives and the guidelines of the task, followed by the list of all possible output labels. The second part of the prompt ensures that the generated outputs remain within the predefined labels set for the given task. Example \ref{ex:promptSD} shows the prompt used for the Stance Detection (SD) task.
The prompts for each task are provided in Appendix \ref{appendix:prompt}.
                
                \noindent
                \begin{mdframed}[backgroundcolor=lightgray!40, roundcorner=10pt]
                    \begin{example}\label{ex:promptSD}

                        \noindent
                        You are an expert in argumentation. Your task is to determine whether the given [SENTENCE] is For or Against. Utilise the [TOPIC] as context to support your decision.
                
                        \noindent
                        Your answer must be in the following format, with only For or Against in the answer section:
                
                        \noindent
                        $<|$ANSWER$|>$ $<$answer$>$ $<|$ANSWER$|>$.\\
    
                        \noindent
                        [TOPIC]: $<$topic$>$
            
                        \noindent
                        [SENTENCE]: $<$sentence$>$              
                    \end{example}
                \end{mdframed}

         All experiments were conducted on an Ubuntu machine equipped with an AMD EPYC 7443 24-core CPU, an NVIDIA A40 48GB GPU, and 65GB of RAM. All fine-tuning performed in this work spanned two epochs with a batch size of 32. We employed Low-Rank Adaptation (LoRA) \citep{hu2022lora} with a rank of 16 to mitigate computational costs and memory requirements while ensuring sufficient parameters to accommodate diverse tasks.

          As the baselines for our evaluation, we employed the Llama3.1-8B-Instruct model in both zero-shot and few-shot settings, along with a DeBERTa model \citep{he2020deberta}. For the few-shot setting, we provided one example per label of the diverse tasks. This resulted in two examples for ACC, CD, ED, and SD, three examples for AR, five for ET, 20 for FD, and three examples for each of the quality dimensions for AQ. 
          
\subsection{Exploiting Model Merging for (Multi-Task) Argument Mining with LLMs}

In this section, we explain how we exploit model merging to create a multi-task AM LLM. The three-steps pipeline (illustrated in Figure \ref{fig:diag_model}) is as follows:

\begin{figure*}[!htb]
\begin{center}
        \includegraphics[width=0.8\linewidth]{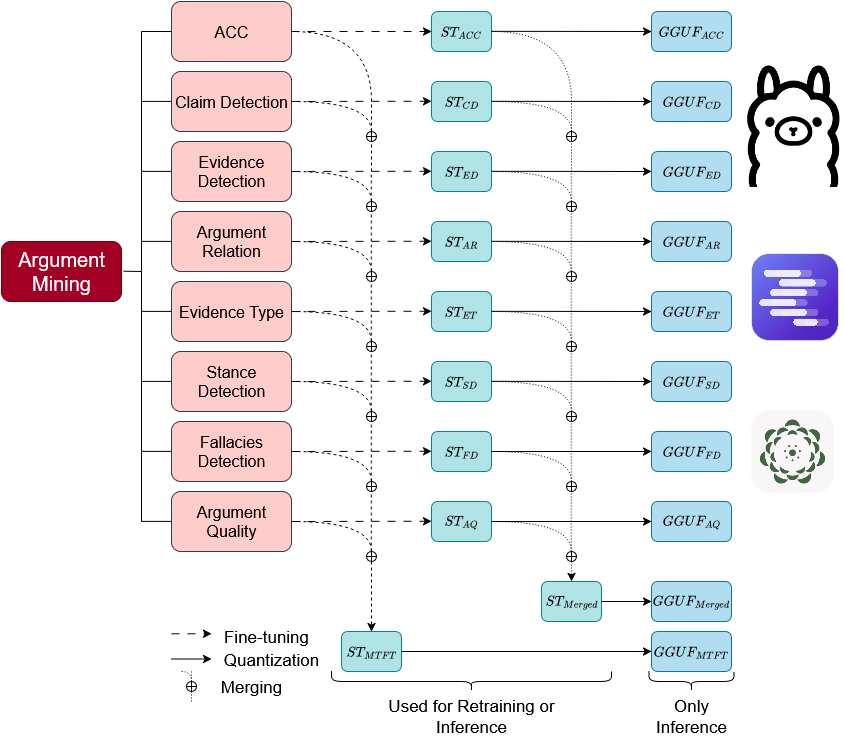}
        \caption{Representation of the merging pipeline. All  fine-tuned models are available in our Hugging Face collection.}
        \label{fig:diag_model}
    \end{center}
\end{figure*}

\begin{enumerate}
    \item We first created a collection of eight models, each fine-tuned for a specific argument mining task, and released them in two formats: Safetensors and GGUF (GGML Universal File). The Safetensors versions enable accurate reproduction of our evaluation results and support potential retraining, while the GGUF format (a compact binary representation that stores both tensors and metadata in a single file) is optimised for efficient local inference on platforms such as Ollama and LM Studio. Both formats are available on Hugging Face\footnote{Our trained LLM collection is accessible at \url{https://huggingface.co/collections/brunoyun/amelia-collection-68518343bf75869b53d0d8bd}}.

    \item 
We categorized the different argumentation tasks into three levels of difficulty: hard, medium, and easy.
This categorization was based on the performance of the various fine-tuned models across the tasks (see Table \ref{table:result_model} in the subsequent section).
Specifically, a task is considered \textit{easy} if most fine-tuned models—including those not specifically trained for that task—achieve over $60\%$ performance. Tasks for which model performance predominantly falls between $50\%$ and $60\%$ are considered \textit{medium}, while tasks with lower performance are classified as \textit{hard}.
Accordingly, the challenging tasks include Fallacies Detection (FD), Argument Quality Assessment (AQ), and Evidence Type Classification (ET). Intermediate tasks comprise Argument Component Classification (ACC) and Argument Relation Classification (AR), whereas the easier tasks are Claim Detection (CD), Evidence Detection (ED), and Stance Detection (SD).
Notably, although most fine-tuned models achieve only around $40\%$ on the AR task, we classify it as medium because merged configurations succeed in obtaining strong performance on this task without retaining a high fraction of task-specific vectors.

\item Subsequently, we merged the eight finely tuned models using the mergekit library \citep{goddard-etal-2024-arcees}. Although there are numerous merging approaches \citep{ilharco2022editing,yadav2023ties,yu2024language,deep2024della}, we evaluated different configurations for two of them: DARE \citep{yu2024language} and DELLA \citep{deep2024della}. On the one hand, the former performs a random pruning on the task vectors, followed by a rescaling method to match the performance of the original model. The latter, on the other hand, enhances DARE using magnitude-based adaptive pruning, which assigns higher probabilities to parameters with larger magnitudes, followed by DARE-like rescaling. This method is designed to preserve significant modifications while minimising interference between task vectors.

We adjusted the following three parameters: the density $\rho$ which represents the fraction of weight to retain in difference to the base model, the weight $w$ of each model task vectors which represents the probability of each model task vector to be used during merging and the $\epsilon$ parameters (used only with the DELLA method) which define the half-width range for the keep probabilities. Keep probabilities for parameters in a task vector will range between $\rho -\epsilon > 0$ and $\rho + \epsilon < 1$.
The configurations for the DELLA method were formulated based on the outcomes of the diverse DARE configurations. The intuition was that pruning the task vectors’ parameters in accordance with their respective importance would enhance the performance of the merging process.
Table \ref{table:param_merge_config} summarises the merging configurations investigated and Table \ref{table:perf_merged_config} shows the performance of the merging methods across the eight different AM tasks.
\end{enumerate}

 \begin{table*}[!htb]
                \begin{center}
                    \resizebox{1\textwidth}{!}{
                        \begin{tabular}{c | c | c | c | c | c | c | c | c | c}
                           \hline
                            Model                                 & \cellcolor{orange!40} ACC                   & \cellcolor{green!40} CD                    & \cellcolor{green!40} ED                    & \cellcolor{orange!40} AR                    & \cellcolor{red!40} ET                    & \cellcolor{green!40} SD                    & \cellcolor{red!40} FD$_{Single}$         & \cellcolor{red!40} FD$_{Multi}$          & \cellcolor{red!40} AQ                    \\
                            \hline
                            Llama-3.1-8B-Instruct zero-shot                & $73.52\%$             & $51.50\%$             & $17.06\%$             & $28.32\%$             & $37.41\%$             & $14.10\%$             & $44.07\%$             & $21.77\%$             & $15.10\%$             \\
                            Llama-3.1-8B-Instruct few-shot                 & $75.47\%$             & $67.83\%$             & $64.20\%$             & $35.97\%$             & $49.31\%$             & $80.00\%$                & $48.50\%$             & $17.25\%$             & $31.83\%$             \\
                            DeBERTa                               & $78.64\%$             & $71.59\%$             & $63.97\%$             & $69.17\%$             & $68.75\%$             & $33.33\%$             & $36.82\%$             & /                     & $48.00\%$                \\
                            \hline
                            \hline
                            Llama-3.1-8B-Instruct FT for ACC       & $\underline{89.61\%}$ & $61.35\%$             & $68.25\%$             & $38.51\%$             & $41.43\%$             & $65.82\%$             & $38.43\%$             & $21.58\%$             & $33.07\%$             \\
                            Llama-3.1-8B-Instruct FT for CD        & $50.18\%$             & $\textbf{85.16\%}$    & $68.91\%$             & $38.29\%$             & $33.91\%$             & $66.97\%$             & $38.90\%$             & $22.67\%$             & $31.24\%$             \\
                            Llama-3.1-8B-Instruct FT for ED        & $63.32\%$             & $74.94\%$             & $\textbf{78.00\%}$       & $28.60\%$             & $38.67\%$             & $68.42\%$             & $39.65\%$             & $18.47\%$             & $29.01\%$             \\
                            Llama-3.1-8B-Instruct FT for AR        & $50.81\%$             & $59.98\%$             & $67.00\%$                & $\underline{87.20\%}$ & $35.07\%$             & $76.00\%$                & $35.14\%$             & $25.86\%$             & $27.97\%$             \\
                            Llama-3.1-8B-Instruct FT for ET        & $56.10\%$             & $67.08\%$             & $61.45\%$             & $26.88\%$             & $\textbf{75.22\%}$    & $69.82\%$             & $46.78\%$             & $29.68\%$             & $29.03\%$             \\
                            Llama-3.1-8B-Instruct FT for SD        & $50.93\%$             & $48.88\%$             & $57.62\%$             & $38.26\%$             & $39.17\%$             & $\underline{94.63\%}$ & $43.23\%$             & $20.99\%$             & $20.39\%$             \\
                            Llama-3.1-8B-Instruct FT for FD        & $66.58\%$             & $65.13\%$             & $64.50\%$             & $38.64\%$             & $46.83\%$             & $64.32\%$             & $\textbf{82.92\%}$    & $50.77\%$             & $41.90\%$             \\
                            Llama-3.1-8B-Instruct FT for AQ        & $74.46\%$             & $59.73\%$             & $68.00\%$                & $30.86\%$             & $44.06\%$             & $60.43\%$             & $47.98\%$             & $24.31\%$             & $\underline{69.54\%}$ \\
                            \hline
                            \hline
                            GGUF$_{ACC}$                          & $87.73\%$             & $63.59\%$             & $63.75\%$             & $36.31\%$             & $37.98\%$             & $64.63\%$             & $30.19\%$             & $29.27\%$             & $32.94\%$             \\
                            GGUF$_{CD}$                           & $54.10\%$             & $81.92\%$             & $60.70\%$             & $36.43\%$             & $31.99\%$             & $63.82\%$             & $30.00\%$                & $31.21\%$             & $33.20\%$             \\
                            GGUF$_{ED}$                           & $56.20\%$             & $63.72\%$             & $71.62\%$             & $34.63\%$             & $36.22\%$             & $61.84\%$             & $34.10\%$             & $34.54\%$             & $34.77\%$             \\
                            GGUF$_{AR}$                           & $55.19\%$             & $60.25\%$             & $63.70\%$             & $84.57\%$             & $31.71\%$             & $76.50\%$             & $29.94\%$             & $34.18\%$             & $32.15\%$             \\
                            GGUF$_{ET}$                           & $58.23\%$             & $64.37\%$             & $58.59\%$             & $29.14\%$             & $72.47\%$             & $68.20\%$             & $39.05\%$             & $32.94\%$             & $31.48\%$             \\
                            GGUF$_{SD}$                           & $56.70\%$             & $50.75\%$             & $57.75\%$             & $38.27\%$             & $33.67\%$             & $93.75\%$             & $34.66\%$             & $30.32\%$             & $21.43\%$             \\
                            GGUF$_{FD}$                           & $62.20\%$             & $59.91\%$             & $62.88\%$             & $35.51\%$             & $42.52\%$             & $64.68\%$             & $74.08\%$             & $\underline{62.16\%}$    & $41.69\%$             \\
                            GGUF$_{AQ}$                           & $67.08\%$             & $59.73\%$             & $69.50\%$             & $31.17\%$             & $41.31\%$             & $61.16\%$             & $41.86\%$             & $30.02\%$             & $66.53\%$             \\
                            \hline
                            Llama-3.1-8B-Instruct FT on Multi-task & $\textbf{90.74\%}$    & $\underline{84.71\%}$ & $\underline{77.75\%}$ & $\textbf{88.33\%}$    & $\underline{73.84\%}$ & $\textbf{95.75\%}$    & $\underline{82.53\%}$ & $50.22\%$             & $\textbf{69.80\%}$    \\
                            GGUF$_{MTFT}$                         & $87.36\%$             & $81.17\%$             & $73.75\%$             & $83.06\%$             & $71.09\%$             & $94.50\%$             & $79.65\%$             & $\textbf{65.32\%}$ & $63.14\%$             \\
                            Merged Model                          & $78.72\%$             & $70.69\%$             & $69.62\%$             & $72.52\%$             & $54.60\%$             & $77.04\%$             & $57.00\%$                & $35.03\%$             & $57.52\%$             \\
                            GGUF$_{Merged}$                       & $65.95\%$             & $65.83\%$             & $62.13\%$             & $62.93\%$             & $49.06\%$             & $74.38\%$             & $50.04\%$             & $40.75\%$             & $44.97\%$             \\
                            \hline
                        \end{tabular}
                    }
                    \caption{Performance of the fine-tuned models merged in 16-bit and the GGUF models in 8-bit.  The cells with \textcolor{red!90}{red} backgrounds indicate the hard tasks, \textcolor{orange}{orange} ones indicate the medium tasks and \textcolor{green}{green} ones indicate the easy tasks.}
                    \label{table:result_model}
                \end{center}
            \end{table*}

 We make the following observations.

                \begin{table*}[!ht]
                    \begin{center}
                        \begin{tabular}{c | c c c | c c c | c c c }
                            \hline
                            Merge configuration     & \multicolumn{3}{c|}{$\rho$} & \multicolumn{3}{c|}{$\epsilon$} & \multicolumn{3}{c}{w} \\
                                               & \cellcolor{red!40} Hard  & \cellcolor{orange!40} Medium & \cellcolor{green!40} Easy   & \cellcolor{red!40} Hard   & \cellcolor{orange!40} Medium & \cellcolor{green!40} Easy  & \cellcolor{red!40} Hard    & \cellcolor{orange!40} Medium & \cellcolor{green!40} Easy     \\
                            \hline
                            DARE \textrm{I}    & $0.5$  & $0.5$ & $0.5$  &        &   /    &       & $0.125$ & $0.125$ & $0.125$ \\
                            DARE \textrm{II}   & $0.7$  & $0.7$ & $0.7$  &        &   /    &       & $0.125$ & $0.125$ & $0.125$ \\
                            DARE \textrm{III}  & $0.85$ & $0.8$ & $0.5$  &        &   /    &       & $0.125$ & $0.125$ & $0.125$ \\
                            DARE \textrm{IV}   & $0.8$  & $0.8$ & $0.8$  &        &   /    &       & $0.2$   & $0.15$  & $0.03$  \\
                            DARE \textrm{V}    & $0.85$ & $0.8$ & $0.5$  &        &   /    &       & $0.2$   & $0.15$  & $0.03$  \\
                            DELLA \textrm{I}   & $0.85$ & $0.8$ & $0.5$  & $0.1$  & $0.1$  & $0.4$ & $0.125$ & $0.125$ & $0.125$ \\
                            DELLA \textrm{II}  & $0.9$  & $0.7$ & $0.5$  & $0.1$  & $0.15$ & $0.4$ & $0.2$   & $0.15$  & $0.03$  \\
                            DELLA \textrm{III} & $0.9$  & $0.85$ & $0.8$ & $0.05$ & $0.1$  & $0.1$ & $0.2$   & $0.15$  & $0.03$  \\
                            \hline
                        \end{tabular}
                        \caption{Parameters for the diverse merging configurations.  The cells with \textcolor{red!90}{red} backgrounds indicate the hard tasks, \textcolor{orange}{orange} ones indicate the medium tasks and \textcolor{green}{green} ones indicate the easy tasks. Within each configuration, tasks with the same difficulty were assigned the same hyperparameter value.}
                        \label{table:param_merge_config}
                    \end{center}
                \end{table*}

\begin{itemize}
    \item  The fine-tuned models on individual tasks consistently outperform the base model in both zero-shot and few-shot settings, as well as the DeBERTa baseline. Notably, the fine-tuned model for the ACC task achieves an $F1$ score of $89.61\%$, a substantial improvement over the zero-shot ($73.52\%$) and few-shot ($75.46\%$) baselines. Similar gains are observed across all tasks, underscoring the effectiveness of task-specific fine-tuning in enhancing model performance.
    
    \item DARE IV and DELLA II achieve high mean F1 scores (61.75\% and 63.64\% respectively). In particular, DELLA II offers the best overall trade-off, with superior performance across multiple categories, including the challenging tasks such as AR and AQ tasks. This suggests that the introduction of non-uniform hyper-parameter values (especially different weights $w$) is beneficial when dealing with heterogeneous task difficulty.

    \item DARE I and II, where all tasks share identical hyperparameters, tend to underperform in terms of mean F1 (59.40\% and 59.18\%). While these models perform consistently, they fail to exploit task-specific adjustments, particularly struggling on harder tasks such as AR, AQ, and ET. This supports the intuition that hard tasks require stronger density and adjusted weight to avoid being overshadowed during merging.

    \item DELLA variants outperform most DARE configurations in mean F1 scores. 
DELLA I and DELLA III perform comparably, but DELLA II benefits the most from balancing relatively high $\rho$ values with selective $\epsilon$ values and non-uniform $w$, achieving the best mean F1. This suggests that carefully tuning can improve generalization on difficult tasks.
    
\end{itemize}

Overall, the DELLA merging method, especially DELLA \textrm{II} and \textrm{III}, are more effective than the DARE configurations. They maintain better performance over different tasks, suggesting that these two configurations are better suited for building a multi-task models for argument mining. When comparing the different models in Table \ref{table:result_model}, ``Merged Model'' refers to our best merged model, i.e., DELLA II.
                
            \begin{table*}[ht]
                \begin{center}
                    \resizebox{\textwidth}{!}{
                        \begin{tabular}{c | c | c | c | c | c | c | c | c | c | c}
                            \hline
                            Merge Configuration     & \cellcolor{orange!40} ACC                & \cellcolor{green!40} CD                 & \cellcolor{green!40} ED                 & \cellcolor{orange!40} AR                 & \cellcolor{red!40} ET                 & \cellcolor{green!40} SD                 & \cellcolor{red!40} FD$_{Single}$      & \cellcolor{red!40} FD$_{Multi}$       & \cellcolor{red!40} AQ                 & Mean $F1$           \\
                            \hline
                            DARE \textrm{I}         & $68.71\%$          & $73.79\%$          & $\textbf{71.88\%}$ & $58.85\%$          & $49.19\%$          & $84.06\%$          & $55.47\%$          & $30.32\%$          & $42.35\%$          & $59.40\%$           \\
                            DARE \textrm{II}        & $68.96\%$          & $74.68\%$          & $71.00\%$             & $59.22\%$          & $49.94\%$          & $83.98\%$          & $53.21\%$          & $30.13\%$          & $41.56\%$          & $59.18\%$           \\
                            DARE \textrm{III}       & $73.09\%$          & $74.31\%$          & $71.75\%$          & $59.60\%$          & $49.69\%$          & $84.35\%$          & $55.52\%$          & $35.03\%$          & $41.30\%$          & $60.51\%$           \\
                            DARE \textrm{IV}        & $\textbf{78.84\%}$ & $69.70\%$          & $69.50\%$          & $68.75\%$          & $54.44\%$          & $74.65\%$          & $57.58\%$          & $35.04\%$          & $51.11\%$          & $62.17\%$           \\
                            DARE \textrm{V}         & $77.71\%$          & $70.45\%$          & $70.00\%$             & $67.75\%$          & $54.19\%$          & $75.68\%$          & $57.00\%$             & $31.55\%$          & $51.50\%$          & $61.75\%$           \\
                            DELLA \textrm{I}        & $74.71\%$          & $\textbf{74.68\%}$ & $71.12\%$          & $63.86\%$          & $50.56\%$          & $\textbf{84.83\%}$ & $57.19\%$          & $\textbf{35.24\%}$ & $44.84\%$          & $61.89\%$           \\
                            DELLA \textrm{II}       & $78.72\%$          & $70.69\%$          & $69.62\%$          & $\textbf{72.52\%}$ & $54.60\%$          & $77.04\%$          & $57.00\%$             & $35.03\%$          & $\textbf{57.52\%}$ & $\textbf{63.64\%}$  \\
                            DELLA \textrm{III}      & $78.60\%$          & $71.70\%$          & $69.37\%$          & $70.64\%$          & $\textbf{54.94\%}$ & $77.20\%$          & $\textbf{58.54\%}$ & $35.03\%$          & $55.16\%$          & $63.46\%$           \\
                            \hline
                        \end{tabular}
                    }
                    \caption{Performance ($F1$ score) of the merge configurations. The cells with \textcolor{red!90}{red} backgrounds indicate the hard tasks, \textcolor{orange}{orange} ones indicate the medium tasks and \textcolor{green}{green} ones indicate the easy tasks.}
                    \label{table:perf_merged_config}
                \end{center}
            \end{table*}
            
\subsection{Fine-tuning a Large Language Model for Multi-Task Argument Mining}

\begin{figure}
                    \centering
                    \includegraphics[width=\linewidth]{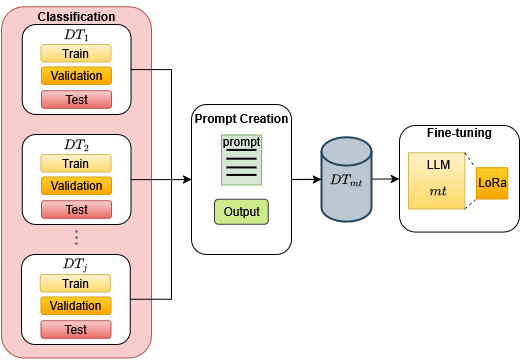}
                    \caption{Representation of the multi-task fine-tuning}
                    \label{fig:mt_ft}
                \end{figure}

We also fine-tuned the Llama 3.1-8B-Instruct model on our multi-task dataset (composed of the training split of all eight AM tasks). This fine-tuned used \texttt{LoRA} with the same parameters as the task-specific fine-tuning (see Section \ref{sec:exp-setting}). 
For the evaluation, we tested the fine-tuned model (referred to as ``Llama-3.1-8B-Instruct FT on Multi-task'' in Table \ref{table:result_model}) on each task's test dataset separately. Figure \ref{fig:mt_ft} shows a representation of the multi-task fine-tuning.
                
This multi-task fine-tuned model achieves the best overall performance, outperforming all other trained models on most tasks. It obtains state-of-the-art results on ACC ($90.74\%$), AR ($88.33\%$), SD ($95.75\%$), and AQ ($69.80\%$), while maintaining comparable performance to the models fine-tuned on individual AM tasks. These results suggest not only an effective transfer of knowledge across tasks—particularly among those with structural similarities such as ACC, CD, ET, and SD—but also that no noticeable conflict arises during joint training, allowing the model to benefit from shared representations without degradation on any individual task.

While the multi-task fine-tuned model achieves superior overall performance compared to the merged model, the latter also offers notable advantages. Model merging provides a flexible and lightweight alternative that does not require joint training on all tasks, making it especially useful when task-specific data is scarce, computational resources are limited, or continual updates are needed. In addition, merging preserves the strengths of task-specialized models without the risk of overfitting to a combined dataset, and allows new tasks to be incorporated incrementally without re-training the entire system. Thus, although multi-task fine-tuning yields the best aggregate results, model merging remains a valuable approach in scenarios where efficiency, modularity, and adaptability are prioritized.

\section{Conclusion}
\label{sect:ccl}

In this work, we introduced AMELIA, a family of (multi-task) end-to-end language models for argument mining. Our contributions are threefold. First, we consolidated and unified 19 widely used argument mining datasets into a common format, thereby providing a large, standardized resource that enables reproducibility and facilitates the application of large language models to diverse argumentative tasks. Second, we conducted an extensive evaluation of fine-tuning strategies using Meta AI’s Llama-3.1-8B-Instruct model, demonstrating that task-specific fine-tuning substantially improves performance across all tasks, while multi-task fine-tuning preserves strong results without degradation, thus confirming the potential of transfer learning across closely related tasks. Third, we explored model merging techniques as a resource-efficient alternative, showing that methods such as DELLA can yield competitive results while maintaining modularity and adaptability.

Our experiments highlight several important insights. 
Multi-task fine-tuning consistently outperforms both zero-shot and few-shot baselines as well as traditional architectures, establishing a new state of the art on multiple tasks. At the same time, model merging offers a practical compromise when computational or data constraints prevent joint training, making it a promising strategy for scalable deployment. Together, these findings underline the flexibility of large language models for argument mining and provide evidence that both fine-tuning and merging can be leveraged to address different application scenarios.

Looking ahead, several promising research directions emerge.
First, while our study focused primarily on classification-based tasks, extending this framework to generative argument mining tasks—such as automatic argument graph construction or debate summarization—would substantially broaden its scope. Pursuing this avenue would require the development of more sophisticated evaluation metrics beyond conventional NLP measures (e.g., ROUGE \cite{lin2004rouge}, METEOR \cite{denkowski2011meteor}, BERTScore \cite{zhang2019bertscore}) and the integration of human or LLM-based evaluations via online platforms to better capture argumentative dimensions that may escape automatic scoring.
Second, incorporating explainability mechanisms and formal argumentation semantics into LLM-based systems would help bridge the gap between high predictive performance and interpretability. This step is critical for deployment in sensitive domains such as law, healthcare, and policy-making, where transparency and accountability are essential.
Third, to further enhance performance, we plan to explore hyperparameter optimisation and prompt engineering. In particular, we will experiment with varying the \texttt{LoRA} rank parameter $r$ (e.g., 32, 64, 128) to identify the most effective configurations across tasks. Additionally, we will investigate optimised prompts in both zero-shot and few-shot settings, with the goal of designing task-specific prompts that maximise model performance.
Finally, we intend to extend our approach to other LLM architectures, thereby enabling a more comprehensive comparison across model families. This includes evaluating instruction-tuned models beyond Llama and systematically analysing performance under consistent experimental settings, in order to better characterise model-specific strengths and limitations in addressing diverse argument mining tasks.

Our work demonstrates that large language models, when carefully adapted through fine-tuning and merging strategies, offer a powerful foundation for advancing argument mining. By releasing the AMELIA models and datasets publicly, we aim to provide the community with both methodological insights and practical tools to accelerate research in computational argumentation.

\section*{Acknowledgments}
This work was carried out as part of the AMELIA project, funded by the Computer Science Department of Université Claude Bernard Lyon 1.

\bibliographystyle{unsrt}  
\bibliography{bibliography}  

\appendix
    \section{All result} \label{appendix: all_res}

    Table \ref{table:all_res_model} includes all results  for the different models studied in this work. $F1$, $P$, and $R$ refer to the F1 score, precision, and recall respectively.
    
              \begin{table*}!hbp
                  \centering
                    \resizebox{1\textwidth}{!}{
                        \begin{tabular}{c | c | c | c | c | c | c | c | c | c}
                           \hline
                            Model                                 & \cellcolor{orange!40} ACC                                                          & \cellcolor{green!40} CD                                                           & \cellcolor{green!40} ED                                                           & \cellcolor{orange!40} AR                                                           & \cellcolor{red!40} ET                                                           & \cellcolor{green!40} SD                                                           & \cellcolor{red!40} FD$_{Single}$                                                & \cellcolor{red!40} FD$_{Multi}$                                              & \cellcolor{red!40} AQ                                                           \\
                                                                  & $F1\qquad P\qquad R$                                         & $F1\qquad P\qquad R$                                         & $F1\qquad P\qquad R$                                         & $F1\qquad P\qquad R$                                         & $F1\qquad P\qquad R$                                         & $F1\qquad P\qquad R$                                         & $F1\qquad P\qquad R$                                         & $F1\qquad P\qquad R$                                      & $F1\qquad P\qquad R$                                         \\
                            \hline
                            Llama 3.1 8B zero-shot                & $0.74\quad0.74\quad0.73$                                     & $0.65\quad0.67\quad0.63$                                     & $0.17\quad0.68\quad0.09$                                     & $0.28\quad0.33\quad0.25$                                     & $0.37\quad0.39\quad0.36$                                     & $0.14\quad0.68\quad0.08$                                     & $0.44\quad0.45\quad0.43$                                     & $0.22\quad0.35\quad0.16$                                  & $0.15\quad0.24\quad0.11$                                     \\
                            Llama 3.1 8B few-shot                 & $0.75\quad0.75\quad0.75$                                     & $0.68\quad0.68\quad0.68$                                     & $0.65\quad0.65\quad0.64$                                     & $0.36\quad0.36\quad0.36$                                     & $0.49\quad0.49\quad0.49$                                     & $0.80\quad0.80\quad0.80$                                     & $0.49\quad0.49\quad0.48$                                     & $0.17\quad0.28\quad0.12$                                  & $0.32\quad0.34\quad0.30$                                     \\
                            DeBERTa                               & $0.79\quad0.79\quad0.79$                                     & $0.72\quad0.72\quad0.72$                                     & $0.64\quad0.64\quad0.64$                                     & $0.69\quad0.69\quad0.69$                                     & $0.69\quad0.73\quad0.70$                                     & $0.33\quad0.25\quad0.5$                                      & $0.37\quad0.44\quad0.38$                                     & /                                                         & $0.48\quad0.47\quad0.55$                                     \\
                            \hline
                            Llama 3.1 8B fine-tuned for ACC       & $\underline{0.90}\quad\underline{0.90}\quad\underline{0.90}$ & $0.61\quad0.61\quad0.61$                                     & $0.68\quad0.68\quad0.68$                                     & $0.39\quad0.39\quad0.39$                                     & $0.41\quad0.41\quad0.41$                                     & $0.66\quad0.67\quad0.65$                                     & $0.38\quad0.39\quad0.38$                                     & $0.21\quad0.33\quad0.15$                                  & $0.33\quad0.33\quad0.33$                                     \\
                            Llama 3.1 8B fine-tuned for CD        & $0.50\quad0.50\quad0.50$                                     & $\textbf{0.85\quad0.85\quad0.85}$                            & $0.70\quad0.70\quad0.70$                                     & $0.38\quad0.38\quad0.38$                                     & $0.34\quad0.34\quad0.34$                                     & $0.67\quad0.68\quad0.66$                                     & $0.39\quad0.40\quad0.38$                                     & $0.23\quad0.40\quad0.16$                                  & $0.31\quad0.31\quad0.31$                                     \\
                            Llama 3.1 8B fine-tuned for ED        & $0.63\quad0.63\quad0.63$                                     & $0.75\quad0.75\quad0.75$                                     & $\textbf{0.78\quad0.78\quad0.78}$                            & $0.29\quad0.29\quad0.29$                                     & $0.39\quad0.39\quad0.39$                                     & $0.68\quad0.70\quad0.66$                                     & $0.40\quad0.40\quad0.39$                                     & $0.18\quad0.27\quad0.14$                                  & $0.29\quad0.29\quad0.29$                                     \\
                            Llama 3.1 8B fine-tuned for AR        & $0.51\quad0.51\quad0.51$                                     & $0.60\quad0.60\quad0.60$                                     & $0.67\quad0.67\quad0.67$                                     & $\underline{0.87}\quad\underline{0.87}\quad\underline{0.87}$ & $0.35\quad0.35\quad0.35$                                     & $0.76\quad0.76\quad0.76$                                     & $0.35\quad0.36\quad0.34$                                     & $0.25\quad0.44\quad0.18$                                  & $0.28\quad0.28\quad0.28$                                     \\
                            Llama 3.1 8B fine-tuned for ET        & $0.56\quad0.56\quad0.56$                                     & $0.67\quad0.67\quad0.67$                                     & $0.61\quad0.61\quad0.61$                                     & $0.27\quad0.27\quad0.27$                                     & $\textbf{0.75\quad0.75\quad0.75}$                            & $0.70\quad0.70\quad0.69$                                     & $0.47\quad0.48\quad0.45$                                     & $0.30\quad0.55\quad0.20$                                  & $0.29\quad0.29\quad0.29$                                     \\
                            Llama 3.1 8B fine-tuned for SD        & $0.51\quad0.51\quad0.51$                                     & $0.49\quad0.49\quad0.49$                                     & $0.58\quad0.58\quad0.58$                                     & $0.39\quad0.39\quad0.39$                                     & $0.39\quad0.39\quad0.39$                                     & $\underline{0.95}\quad\underline{0.95}\quad\underline{0.95}$ & $0.43\quad0.43\quad0.43$                                     & $0.21\quad0.33\quad0.15$                                  & $0.20\quad0.20\quad0.20$                                     \\
                            Llama 3.1 8B fine-tuned for FD        & $0.67\quad0.67\quad0.67$                                     & $0.65\quad0.65\quad0.65$                                     & $0.65\quad0.65\quad0.65$                                     & $0.39\quad0.39\quad0.39$                                     & $0.47\quad0.47\quad0.47$                                     & $0.64\quad0.70\quad0.59$                                     & $\textbf{0.83\quad0.83\quad0.83}$                            & $0.51\quad\textbf{0.83}\quad0.34$             & $0.42\quad0.42\quad0.42$                                     \\
                            Llama 3.1 8B fine-tuned for AQ        & $0.74\quad0.74\quad0.74$                                     & $0.60\quad0.60\quad0.60$                                     & $0.60\quad0.60\quad0.60$                                     & $0.31\quad0.31\quad0.31$                                     & $0.44\quad0.44\quad0.44$                                     & $0.60\quad0.68\quad0.54$                                     & $0.48\quad0.48\quad0.48$                                     & $0.24\quad0.40\quad0.18$                                  & $\underline{0.70}\quad\underline{0.70}\quad\underline{0.70}$ \\
                            \hline
                            \hline
                            GGUF$_{ACC}$                          & $0.88\quad0.88\quad0.88$                                     & $0.64\quad0.64\quad0.64$                                     & $0.64\quad0.64\quad0.64$                                     & $0.36\quad0.36\quad0.36$                                     & $0.38\quad0.38\quad0.38$                                     & $0.65\quad0.65\quad0.65$                                     & $0.30\quad0.32\quad0.29$                                     & $0.29\quad0.30\quad0.28$                                  & $0.33\quad0.33\quad0.33$                                     \\
                            GGUF$_{CD}$                           & $0.54\quad0.54\quad0.54$                                     & $0.82\quad0.82\quad0.82$                                     & $0.61\quad0.63\quad0.59$                                     & $0.36\quad0.36\quad0.36$                                     & $0.32\quad0.32\quad0.32$                                     & $0.64\quad0.65\quad0.63$                                     & $0.30\quad0.31\quad0.29$                                     & $0.31\quad0.32\quad0.30$                                  & $0.33\quad0.33\quad0.33$                                     \\
                            GGUF$_{ED}$                           & $0.56\quad0.56\quad0.56$                                     & $0.64\quad0.64\quad0.64$                                     & $0.72\quad0.72\quad0.72$                                     & $0.35\quad0.35\quad0.35$                                     & $0.36\quad0.36\quad0.36$                                     & $0.62\quad0.64\quad0.60$                                     & $0.34\quad0.37\quad0.31$                                     & $0.35\quad0.36\quad0.33$                                  & $0.35\quad0.35\quad0.35$                                     \\
                            GGUF$_{AR}$                           & $0.55\quad0.55\quad0.55$                                     & $0.60\quad0.60\quad0.60$                                     & $0.64\quad0.64\quad0.64$                                     & $0.85\quad0.85\quad0.85$                                     & $0.32\quad0.32\quad0.32$                                     & $0.77\quad0.77\quad0.77$                                     & $0.30\quad0.31\quad0.29$                                     & $0.34\quad0.36\quad0.32$                                  & $0.32\quad0.32\quad0.32$                                     \\
                            GGUF$_{ET}$                           & $0.58\quad0.59\quad0.57$                                     & $0.64\quad0.64\quad0.64$                                     & $0.59\quad0.59\quad0.59$                                     & $0.29\quad0.31\quad0.27$                                     & $0.72\quad0.72\quad0.72$                                     & $0.68\quad0.70\quad0.66$                                     & $0.39\quad0.40\quad0.38$                                     & $0.33\quad0.41\quad0.28$                                  & $0.31\quad0.31\quad0.31$                                     \\
                            GGUF$_{SD}$                           & $0.57\quad0.57\quad0.57$                                     & $0.51\quad0.51\quad0.51$                                     & $0.58\quad0.58\quad0.58$                                     & $0.38\quad0.38\quad0.38$                                     & $0.34\quad0.34\quad0.34$                                     & $0.94\quad0.94\quad0.94$                                     & $0.35\quad0.37\quad0.33$                                     & $0.30\quad0.31\quad0.29$                                  & $0.21\quad0.21\quad0.21$                                     \\
                            GGUF$_{FD}$                           & $0.62\quad0.62\quad0.62$                                     & $0.60\quad0.62\quad0.58$                                     & $0.63\quad0.63\quad0.63$                                     & $0.36\quad0.36\quad0.36$                                     & $0.43\quad0.43\quad0.43$                                     & $0.65\quad0.71\quad0.60$                                     & $0.74\quad0.74\quad0.74$                                     & $\underline{0.62}\quad0.74\quad\textbf{0.53}$             & $0.42\quad0.42\quad0.42$                                     \\
                            GGUF$_{AQ}$                           & $0.67\quad0.67\quad0.67$                                     & $0.60\quad0.60\quad0.60$                                     & $0.70\quad0.70\quad0.70$                                     & $0.31\quad0.31\quad0.31$                                     & $0.41\quad0.41\quad0.41$                                     & $0.61\quad0.69\quad0.55$                                     & $0.42\quad0.42\quad0.42$                                     & $0.30\quad0.33\quad0.27$                                  & $0.67\quad0.67\quad0.67$                                     \\
                            \hline
                            Llama 3.1 8B fine-tuned on Multi-task & $\textbf{0.91\quad0.91\quad0.91}$                            & $\underline{0.85}\quad\underline{0.85}\quad\underline{0.85}$ & $\underline{0.78}\quad\underline{0.78}\quad\underline{0.78}$ & $\textbf{0.88\quad0.88\quad0.88}$                            & $\underline{0.74}\quad\underline{0.74}\quad\underline{0.74}$ & $\textbf{0.96\quad0.96\quad0.96}$                            & $\underline{0.83}\quad\underline{0.83}\quad\underline{0.83}$ & $0.50\quad\underline{0.83}\quad0.36$          & $\textbf{0.70\quad0.70\quad0.70}$                            \\
                            GGUF$_{MTFT}$                         & $0.87\quad0.87\quad0.87$                                     & $0.81\quad0.81\quad0.81$                                     & $0.74\quad0.74\quad0.74$                                     & $0.83\quad0.83\quad0.83$                                     & $0.71\quad0.71\quad0.71$                                     & $0.94\quad0.95\quad0.94$                                     & $0.80\quad0.80\quad0.80$                                     & $\textbf{0.65}\quad0.83\quad\underline{0.53}$             & $0.63\quad0.63\quad0.63$                                     \\
                            ST$_{Merged}$                         & $0.78\quad0.78\quad0.78$                                     & $0.71\quad0.71\quad0.71$                                     & $0.70\quad0.70\quad0.70$                                     & $0.73\quad0.73\quad0.73$                                     & $0.55\quad0.55\quad0.55$                                     & $0.77\quad0.78\quad0.76$                                     & $0.57\quad0.57\quad0.57$                                     & $0.35\quad0.60\quad0.25$                                  & $0.58\quad0.58\quad0.58$                                     \\
                            GGUF$_{Merged}$                       & $0.66\quad0.66\quad0.66$                                     & $0.66\quad0.66\quad0.66$                                     & $0.62\quad0.62\quad0.62$                                     & $0.63\quad0.63\quad0.63$                                     & $0.49\quad0.49\quad0.49$                                     & $0.74\quad0.75\quad0.73$                                     & $0.50\quad0.50\quad0.50$                                     & $0.41\quad0.45\quad0.37$                                  & $0.45\quad0.45\quad0.45$                                     \\
                            \hline
                        \end{tabular}
                    }
                    \caption{Performance (F1, Precision, Recall) of the models merged in 16 bit and the GGUF model in 8 bit}
                    \label{table:all_res_model}
                
            \end{table*}
    \section{Prompts} \label{appendix:prompt}
        \subsection{Argument component classification}
        \begin{mdframed}[backgroundcolor=lightgray!40, roundcorner=10pt]
            You are an expert in argumentation. Your task is to determine whether the given [SENTENCE] is a Claim or a Premise. Utilize the [TOPIC] and the [FULL TEXT] as context to support your decision.
            
            \noindent
            Your answer must be in the following format with only Claim or Premise in the answer section:
            
            \noindent
            $<|$ANSWER$|>$ $<$answer$>$ $<|$ANSWER$|>$.\\

            \noindent
            [TOPIC]: $<$topic$>$

            \noindent
            [SENTENCE]: $<$sentence$>$

            \noindent
            [FULL TEXT]: $<$full text$>$
            \end{mdframed}
        \subsection{Claim detection}
        \begin{mdframed}[backgroundcolor=lightgray!40, roundcorner=10pt]
            You are an expert in argumentation. Your task is to determine whether the given [SENTENCE] is a Claim or Non-claim. Utilize the [TOPIC] and the [FULL TEXT] as context to support your decision.
            
            \noindent
            Your answer must be in the following format with only Claim or Non-claim in the answer section:
            
            \noindent
            $<|$ANSWER$|>$ $<$answer$>$ $<|$ANSWER$|>$.\\

            \noindent
            [TOPIC]: $<$topic$>$ 

            \noindent
            [SENTENCE]: $<$sentence$>$

            \noindent
            [FULL TEXT]: $<$full text$>$
            \end{mdframed}
        \subsection{Evidence detection}
        \begin{mdframed}[backgroundcolor=lightgray!40, roundcorner=10pt]
            You are an expert in argumentation. Your task is to determine whether the given [SENTENCE] is an Evidence or Non-evidence. Utilize the [TOPIC] and the [ARGUMENT] as context to support your decision.
            
            \noindent
            Your answer must be in the following format with only Evidence or Non-evidence in the answer section:
            
            \noindent
            $<|$ANSWER$|>$ $<$answer$>$ $<|$ANSWER$|>$.\\

            \noindent
            [TOPIC]: $<$topic$>$

            \noindent
            [ARGUMENT]: $<$argument$>$

            \noindent
            [SENTENCE]: $<$sentence$>$
            \end{mdframed}
        \subsection{Argument relation}
        \begin{mdframed}[backgroundcolor=lightgray!40, roundcorner=10pt]
            You are an expert in argumentation. Your task is to determine the type of relation between [SOURCE] and [TARGET]. The type of relation would be in the [RELATION] set. Utilize the [TOPIC] as context to support your decision.
            
            \noindent
            Your answer must be in the following format with only the type of the relation in the answer section:
            
            \noindent
            $<|$ANSWER$|>$ $<$answer$>$ $<|$ANSWER$|>$.\\

            \noindent
            [RELATION]: $\{$relation$\}$

            \noindent
            [SOURCE]: $<$source argument$>$

            \noindent
            [TARGET]: $<$argument target$>$
            \end{mdframed}
        \subsection{Evidence type}
        \begin{mdframed}[backgroundcolor=lightgray!40, roundcorner=10pt]
            You are an expert in argumentation. Your task is to determine the type of evidence of the given [SENTENCE]. The type of evidence would be in the [TYPE] set. Utilize the [TOPIC] and the [CLAIM] as context to support your decision.
            
            \noindent
            Your answer must be in the following format with only the type of evidence in the answer section:
            
            \noindent
            $<|$ANSWER$|>$ $<$answer$>$ $<|$ANSWER$|>$.\\

            \noindent
            [TYPE]: $\{$type$\}$

            \noindent
            [TOPIC]: $<$topic$>$

            \noindent
            [SENTENCE]: $<$sentence$>$
            \end{mdframed}
        \subsection{Stance detection}
        \begin{mdframed}[backgroundcolor=lightgray!40, roundcorner=10pt]
            You are an expert in argumentation. Your task is to determine whether the given [SENTENCE] is For or Against. Utilize the [TOPIC] as context to support your decision.
            
            \noindent
            Your answer must be in the following format with only For or Against in the answer section:
            
            \noindent
            $<|$ANSWER$|>$ $<$answer$>$ $<|$ANSWER$|>$.\\

            \noindent
            [TOPIC]: $<$topic$>$

            \noindent
            [SENTENCE]: $<$sentence$>$
            \end{mdframed}
        \subsection{Fallacies detection}
        \begin{mdframed}[backgroundcolor=lightgray!40, roundcorner=10pt]
            You are an expert in argumentation. Your task is to determine the type of fallacy in the given [SENTENCE]. The fallacy would be in the [FALLACY] Set. Utilize the [TITLE] and the [FULL TEXT] as context to support your decision.
            
            \noindent
            Your answer must be in the following format with only the fallacy in the answer section:
            
            \noindent
            $<|$ANSWER$|>$ $<$answer$>$ $<|$ANSWER$|>$.\\

            \noindent
            [FALLACY]: $\{$fallacies$\}$

            \noindent
            [TITLE]: $<$title$>$

            \noindent
            [SENTENCE]: $<$sentence$>$

            \noindent
            [FULL TEXT]: $<$full text$>$
            \end{mdframed}
        \subsection{Argument quality assessment}
        \begin{mdframed}[backgroundcolor=lightgray!40, roundcorner=10pt]
            You are an expert in argumentation. Your task is to determine the quality of the [SENTENCE]. The quality would be in the [QUALITY] set. Utilize the [TOPIC], the [STANCE] and the [DEFINITION] as context to support your decision.
            
            \noindent
            Your answer must be in the following format with only the quality in the answer section:
            
            \noindent
            $<|$ANSWER$|>$ $<$answer$>$ $<|$ANSWER$|>$.\\

            \noindent
            [QUALITY]: $\{$quality$\}$

            \noindent
            [TOPIC]: $<$topic$>$

            \noindent
            [STANCE]: $<$stance$>$

            \noindent
            [DEFINITION]: $<$definition$>$

            \noindent
            [SENTENCE]: $<$sentence$>$
            \end{mdframed}
\end{document}